\documentclass[]{article}
\usepackage[letterpaper]{geometry}
\usepackage{amta2022}
\usepackage{times}
\usepackage{url}
\usepackage{latexsym}
\usepackage{natbib}
\usepackage{layout}

\usepackage{graphics,caption,tabto,tabularx,multicol,float,wrapfig}
\usepackage[hidelinks]{hyperref}
\usepackage[all]{nowidow}

\begin{document}

\title{\bf A Snapshot into the Possibility of Video Game Machine Translation}

\author{\name{\bf Damien Hansen~$^{1,2,3}$}\hfill\addr{damien.hansen@uliege.be}\\
        \name{\bf Pierre-Yves Houlmont~$^{1,2}$}\hfill\addr{pyhoulmont@uliege.be}\\
        \addr{$^1$~University of Liège, CIRTI*, 4020 Liège, Belgium}\\
        \addr{$^2$~University of Liège, Liège Game Lab, 4000 Liège, Belgium}\\
        \addr{$^3$~Univ. Grenoble Alpes, CNRS, Grenoble INP**, LIG, 38000 Grenoble, France}}

\maketitle
\pagestyle{empty}

\begin{abstract}
    We present in this article what we believe to be one of the first attempts at video game machine translation. Our study shows that models trained only with limited in-domain data surpass publicly available systems by a significant margin, and a subsequent human evaluation reveals interesting findings in the final translation. The first part of the article introduces some of the challenges of video game translation, some of the existing literature, as well as the systems and data sets used in this experiment. The last sections discuss our analysis of the resulting translation and the potential benefits of such an automated system. One such finding highlights the model's ability to learn typical rules and patterns of video game translations from English into French. Our conclusions therefore indicate that the specific case of video game machine translation could prove very much useful given the encouraging results, the highly repetitive nature of the work, and the often poor working conditions that translators face in this field. As with other use cases of MT in cultural sectors, however, we believe this is heavily dependent on the proper implementation of the tool, which should be used interactively by human translators to stimulate creativity instead of raw post-editing for the sake of productivity.
\end{abstract}

\renewcommand{\thefootnote}{\fnsymbol{footnote}}
\footnotetext[1]{Centre Interdisciplinaire de Recherche en Traduction et en Interprétation.}
\footnotetext[7]{Institute of Engineering Univ. Grenoble Alpes.}
\renewcommand{\thefootnote}{\arabic{footnote}}

\vspace{-0.5em}\section{Introduction}

Since the apparition of recurrent neural networks with attention mechanisms \citep{bahdanau2014}, neural machine translation (NMT) has improved to the point of becoming the default paradigm for this task. With new architectures such as the Transformer \citep{vaswani2017} and a similarly growing number of domain adaptation techniques \citep{chu2018}, NMT has also started being used in increasingly more complex domains, and even tailored to the production of specific translators and companies.

The video game market similarly seems to have been ever growing in the last decades, becoming one of the fastest growing and highest grossing industries today within the cultural and entertainment sectors. The recent global health crisis further reinforced this trend and showed that games have an important role to play beyond just entertainment, as can be seen in a recent EU report on the cultural and creative sectors (CCS) in Europe \citep{IDEA2021}: ``Few winners can be found in the CCS during the COVID-19 global crisis. One of those, together with streaming platforms, is the video games sub-sector. With a turnover of EUR 21.6 billion and a 3\% year-on-year growth from 2018, the gaming sub-sector has proven to be strong also in hard economic times.''

As part of this success, it has now been expected for a while that video games be translated not just into the Western E-FIGS (English to French, Italian, German and Spanish), but into at least eight or ten languages in order to be considered a profitable and successful enterprise \citep{bernalmerino2015}. This creates, in turn, a huge demand for translation and a challenge in its own right, especially for smaller or newly formed studios.

It should furthermore be noted that, nowadays, video game localization is mainly considered from a market perspective. Indeed, the main purpose of this process is to make a given title available to a larger audience. To do so, several attempts have been made to produce game localization manifestos encouraging best practices \citep{chandlerdeming2012,honeywoodfung2012}.

Those practices --- and scientific studies in the field of video game localization in general --- are said to be market-driven \citep{ohagan2013}. Even though video games represent an \mbox{exceptional} place of intercultural contact, cultural traces are quasi systematically neutralized in an attempt to ensure sales, to the point where they would be considered as localization errors were they to remain in the target version \citep{mandiberg2015}, despite evidence from reception studies showing that current localization practices do not correspond to what players expect from a cultural point of view \citep{ellefsenbernalmerino2018,fernandezcostales2016,geurts2015}.

On the other hand, the video game industry, as is the case with other digital productions\footnote{~As the Web gave rise to the development of a ``user-participatory culture'' \citep{jenkins2006} and ``user-generated content'' \citep{vandijk2009}, the translation sector similarly witnessed the emergence of ``user'' or ``amateur'' crowd-sourced translations \citep{lavaultolleon2011,gambier2016,doherty2016}.}, is characterized by a particularly high level of fan-made content \mbox{\citep{barnabe2015,hurel2022}}, with undertakings varying from patches\footnote{~A patch, or fix, is a light modification made to a program in order to correct bugs, for example, to add a translation or to improve its usability.}, mods\footnote{~Mods affect the game more deeply and generally aim at customizing or expanding rather than fixing it. These modifications can affect its appearance, its story or even its gameplay \citep{barnabe2015}.}, or even (re)creations that far outgrow the original games themselves. It is also very common for fans and volunteers to look into the original translations, sometimes correcting minor mistakes in these very large works, sometimes coming up with entirely new localization projects if the title was never translated or if the translation was not done by professionals and its quality deemed too poor \citep{diazmonton2007,munozsanchez2009,vazquezcalvo2019}.

Indeed, translation has traditionally been seen in the market as a trivial and disposable process that could be targeted to cut costs, leading up to the numerous examples of bad video game translations that careless uses of machine translation (MT) definitely did not help with \citep{mandelin2017}. This brings several issues, not only for the players, but for the companies as well. As \cite{bernalmerino2008} explains: ``One of the most common complaints we can read about in internet forums is the lack of translation quality, and how sometimes they have to go back to the original version to find out what to do and how. In many cases, even though the game might have been translated and it is playable, fans fail to be impressed because the poor quality of the localisation defeats its own purpose: to thrill and engage the gamer. [\ldots] Language and culture are ever-present elements in us and the things we do, players cannot help but notice continuous serious mistakes in the game, and it will erode their trust in some developer and publisher brands.''

But what makes video game translation such an arduous task and how could machines play a role in it if it already proves challenging for humans? In addition to the common pitfalls of translation and creative works, translators in the field are typically presented with spreadsheets in which text segments are not ordered chronologically and do not offer any other kind of contextual information, concerning for instance the communication medium, the type of \mbox{discourse}, the person speaking, etc. \citep{diazmonton2007}. What is more, a survey conducted with translators working in this sector found that only 30\% of them had access to the game they were asked to translate \citep{theroine2021}. The localization of video games, however, is dependent on more than just text, as it relies on a precise balancing of intersemiotic dynamics. Indeed, the different semiotic systems of a given game work together to provide a complete and coherent media experience, and may present ``tensions'' or a form of ``interdependency'' that can limit the spectrum of possibilities for the translator \citep{houlmont2022}. This lack of audiovisual or contextual cues and text linearity is so prevalent that it has been theorized by \cite{bernalmerino2015} as a double-blind process. In a sense, machines would therefore have as much \mbox{(or rather as little)} information at their disposal as humans do when translating video games. This observation thus makes us wonder what their performance would be in this scenario, but more importantly how useful they could be to translators.

\section{State of the Art}

As we mentioned, the advances in NMT architectures and domain adaptation techniques allowed for better performances in various domains. In the cultural sector, machine translation has mostly been used in the process of film subtitling, but a growing number of studies has focused on the challenges of its development, for instance, with literary texts \citep{sakamoto2020,hansen2021}.

Concerning video games, it is hard to find scientific sources that go further than acknowledging the existence of MT within this sector, as in \cite{mangiron2013}. This use of machine translation is further confirmed in various online posts and articles, but also in sessions dedicated to localization at the Game Developers Conference \citep{bartlet2014,IGDA2018}. Some online translation service providers also put it forward, alongside with the traditional arguments of cost and time savings that typically accompany MT, and it would appear that some studios and translators are sometimes forced to turn to this technology in order to meet the strict and tight deadlines of the industry. It remains nevertheless agreed that MT would never be suitable for the translation of the more creative in-game contents, such as dialogues \citep{mangiron2013}. We can once again find confirmation for this observation in online interviews \citep{ruete2021}, but also in the deployment and user feedback of open source tools like \textit{RetroArch} (Libretro, 2019), that leverage Google’s translation service to make retro games accessible to a larger public.

Machine translation, however, might benefit the video game localization industry, whether it is to cope with the heavy requirements of a simultaneous multilingual game delivery, to help fans make their content available to more people, or to assist professionals during this long-term and demanding task. Yet, we have found no evidence of MT tools being specifically adapted for the translation of video games, and the existing use of MT as well as the advances in NMT make it the perfect time to try and see how effective such a custom engine could be. To our knowledge, this is thus the first article to dive deeply into the possibility of video game machine translation, or, at the very least, that an MT system was specifically designed for such a task.

\section{Methodology}

Video game translation is an eminently difficult task that involves linguistic skills and cultural awareness from the two languages at hand, diegetic knowledge about the world in which the game takes place, as well as an understanding of how the game is to be played in real life and how the translation fits into its development. To have an idea of whether MT could be of help to professional translators, we built a first system tailored to the translation of video games for the English–French language pair.

This system was trained with the 2nd edition of the OpenNMT framework \citep{klein2017} on a custom-made corpus of video game translations. Our bespoke MT engine was then evaluated on two role-playing games (RPG), namely \textit{The Elder Scrolls V: Skyrim} (Bethesda Game Studios, 2011) and \textit{Fallout 4} (Bethesda Game Studios, 2015), taking place in a high fantasy and post-apocalyptic setting respectively.

\subsection{Data set}

For this task, we used a relatively small in-domain data set (of about 1 million sentences), that combines the official translations for 22 games, mainly from the fantasy and post-apocalyptic RPG genre.\footnote{~For a more detailed overview of the corpus, see appendix A.} The same corpus was used to train both models, but we extracted validation and test samples without replacement for each game being evaluated.

\begin{multicols}{2}
\renewcommand{\arraystretch}{1.3}
\begin{table}[H]
\captionsetup{justification=centering}
\noindent\resizebox{\columnwidth}{!}{
\begin{tabular}{|l|r|r|r|}\cline{2-4}
    \multicolumn{1}{l|}{} & \textbf{Sentences} & \textbf{EN tokens} & \textbf{FR tokens} \\\hline
    \textbf{Training}     & 951,373            & 16,942,551         & 18,008,101         \\\hline
    \textbf{Validation}   & 4,785              & 70,576             & 74,189             \\\hline
    \textbf{Test}         & 501                & 8,574              & 9,254              \\\hline
\end{tabular}}
\caption{Data sets for \textit{The Elder Scrolls V: Skyrim} (Bethesda Game Studios, 2011).}
\label{table1}
\end{table}

\begin{table}[H]
\captionsetup{justification=centering}
\noindent\resizebox{\columnwidth}{!}{
\begin{tabular}{|l|r|r|r|}\cline{2-4}
    \multicolumn{1}{l|}{} & \textbf{Sentences} & \textbf{EN tokens} & \textbf{FR tokens} \\\hline
    \textbf{Training}     & 951,861            & 16,958,940         & 18,020,052         \\\hline
    \textbf{Validation}   & 4,401              & 56,486             & 64,738             \\\hline
    \textbf{Test}         & 397                & 6,275              & 6,754              \\\hline
\end{tabular}}
\caption{Data sets for \textit{Fallout 4}\\(Bethesda Game Studios, 2015).}
\label{table2}
\end{table}
\end{multicols}

\vspace{-1em}

These titles were chosen not only for their wide popularity, but also because they are representative of the multiple difficulties professionals might have to face in this industry. One of them is the sheer volume of in-game content to be translated, which often results in multiple translators working on the same project \citep{diazmonton2007}. On top of this, both games make use of a very specific terminology that draws from their distinct fictional universe. This abundance of lexical creations, or \textit{irrealia} \citep{loponen2009}, is a challenge in its own right, especially if the work is collaborative in nature or if professionals are not familiar with it, but this aspect is all the more important as this terminology builds on previous titles.

The Elder Scrolls and the Fallout series have indeed distinguished themselves through their rich mythopoetic universe, which partly explains why fans have taken such an interest in creating content in and around the games. Proof of this is the breadth of the fictional literature found in \textit{Skyrim} for instance. The number of these metadiegetic works and their size --- several hundred pages for some --- actually lead us to delete all books from the Elder Scrolls games in our corpus, as it would have needed an additional and separate alignment process, but they were the only material removed from the original files.

The acquisition method for this corpus was inspired by the work of hobbyist modders. This type of content is a useful way to foster player engagement and to extend the lifespan of a game, which is why developers and distributors facilitate the creation and sharing of content through the release of toolkits or dedicated interfaces such as the workshop section in \textit{Steam} (Valve, 2003). Today, platforms such as the mod repository \textit{Nexus Mods}\footnote{~\url{https://www.nexusmods.com/}.} or the amateur French translation team \textit{La Confrérie des Traducteurs}\footnote{~\url{https://www.confrerie-des-traducteurs.fr/}.} serve as a testament to the dedication of some of the fans. Among these mods, a few pieces of software are built from scratch to help with the translation process.\footnote{~It might be important to keep in mind here that such practices are not at all intended to replace professional translators, but only to extend the reach of works that would never be translated otherwise.} They rely on custom translation memories (TMs) and allow the user to load translation files from existing games to help speed up and improve \mbox{the quality of the translation.} And although most of these users are probably not even aware of this, the process is exactly similar to that of computer-assisted translation tools, which have long been used by professionals.\footnote{~Apart from in-house teams however, the use of TMs remains relatively uncommon in the video game industry.}

In a similar fashion, we were able to retrieve the translation files for a number of games, convert them from their native format into spreadsheet files, and align each pair in a standard translation memory format though the \textit{Heartsome TMX Editor 8.0} (Heartsome Technologies Ltd, 2014). The resulting bitexts were cleaned semi-manually with the same program and a few regex manipulations to discard duplicate, empty and untranslated segments, to normalize the extremely varied typographic conventions and to remove game-specific formatting --- most tags and variables were still left as such. We then extracted and compiled the remaining 956,661 segments, many of which contain more than one sentence.

Since 1 million bilingual segments is scarce for NMT --- 6 million is even considered ``frugal'' by today's standards \citep{blin2021} --- we also built models which we fine-tuned on our gaming data set. The initial training was performed on common generic data sets for the same language pair: \textit{Books}, \textit{Europarl-v8}, \textit{GlobalVoices-2018}, \textit{News-Commentary-v16}, and \textit{TED2020} \citep{tiedemann2012}, for a total of 2,923,826 additional aligned sentences.\footnote{~An experiment was also carried out with the dozens of millions of sentences from the WMT 2014 translation task \citep{bojar2014}, but the final score dropped systematically due to the quality and dissimilarity of these data sets.}

\subsection{Architecture}

In each case, data sets were tokenized with Moses \citep{koehn2007} and further segmented with sentencepiece's unigram model \citep{kudo2018}, using a vocabulary size of 32,000 tokens. We then trained two Transformer models for each scenario (fine-tuned and in-domain only) with OpenNMT's default parameters and the ``base'' architecture described in \cite{vaswani2017}: 6~encoder and decoder layers, 8~attention heads, a dimension of 512 and feed-forward layers of size 2048, with dropout 0.1. We stopped training models after convergence (200,000~steps) and translated all files with the default parameters of OpenNMT again, which uses a beam size of 5 at inference.

\section{Results}

For evaluation, we detokenized the output with the same Moses script and computed a BLEU score with sacreBLEU \citep{post2018}. We first report in Table~\ref{table3} the score obtained on \textit{The Elder Scrolls~V: Skyrim} and \textit{Fallout~4}, two games for which we could compare the output with the official translation. These test sets contained mostly dialogues and, to a lesser extent, quest descriptions, action choices and instructions.

\begin{wraptable}{r}{.47\textwidth}
\begin{center}\captionsetup{margin=0.5em}
\begin{tabular}{|l|c|c|}\cline{2-3}
    \multicolumn{1}{l|}{} & \textbf{In-domain} & \textbf{Fine-tuned} \\\hline
    \textbf{Skyrim}       & 37.14              & 37.38               \\\hline
    \textbf{Fallout 4}    & 31.18              & 30.52               \\\hline
\end{tabular}
\caption{Score given by sacreBLEU for the in-domain only and fine-tuned models on both games.}
\label{table3}
\end{center}
\end{wraptable}

Our first comparison indicated that adding data did not necessarily increase performance. In the first case, the added 3 million \mbox{segments} resulted in a negligible increase in BLEU, whereas that score even decreased in the second case. For this reason, and since our aim was to evaluate systems trained on in-domain data, we discarded the tuned models. This decision was further motivated by our evaluation of the resulting translations, which showed that the in-domain models seemed to learn inherent rules of the domain which the fine-tuned models did not, such as the fact that gendered variables tend to disappear when translating into French:

{\small\setlength{\parskip}{0.6em}\leftskip=1.2em

\noindent\textbf{SRC:} Is it true what they say? There was a dragon held captive in Whiterun, and you... you released it? \textbf{By the gods, woman, why?}

\noindent\textbf{HYP (fine-tuned):} C'est vrai ce qu'ils disent~? Il y avait un dragon captif à Blancherive, et vous... vous l'avez libéré~? \textbf{Par les dieux, femme, pourquoi~?}

\noindent\textbf{HYP (in-domain):} C'est vrai ce qu'on raconte~? Il y avait un dragon emprisonné à Blancherive et vous... vous l'avez libéré~? \textbf{Par les dieux, pourquoi~?}

\noindent\textbf{REF:} C'est vrai, ce qu'on raconte~? Qu'il y avait un dragon captif à Blancherive et que vous l'avez délivré~? \textbf{Mais pourquoi, par les dieux~?}

}~\vspace{-.4em}

\noindent In doing so, we could assess more accurately the performance of systems trained only on data from the video game domain, which are already very encouraging considering that we used such a small training data set. To illustrate these results more clearly, we provide in Table~\ref{table4} and Table~\ref{table5} a comparison with publicly available systems using three measures provided by sacreBLEU. For each, an arrow indicates if the improvement is reflected by a higher or a lower score.\footnote{~Metric signatures for sacreBLEU (publicly available systems tested on 23/12/2020):\\\tabto{2.1em}BLEU\tabto{6em}\#:1$\vert$c:mixed$\vert$e:no$\vert$tok:13a$\vert$s:exp$\vert$v:2.0.0\\\tabto{2.1em}chrF2++\tabto{6em}\#:1$\vert$c:mixed$\vert$e:yes$\vert$nc:6$\vert$nw:2$\vert$s:no$\vert$v:2.0.0\\\tabto{2.1em}TER\tabto{6em}\#:1$\vert$c:lc$\vert$t:tercom$\vert$nr:no$\vert$pn:yes$\vert$a:no$\vert$v:2.0.0.}

\begin{multicols}{2}
\begin{table}[H]
\renewcommand{\arraystretch}{1.3}
\captionsetup{justification=centering}
\noindent\resizebox{\columnwidth}{!}{
\begin{tabular}{|l|c|c|c|}\cline{2-4}
    \multicolumn{1}{l|}{} & \textbf{BLEU $\Uparrow$} & \textbf{chrF2++ $\Uparrow$} & \textbf{TER $\Downarrow$} \\\hline
    \textbf{Google Translate} & 27.75 & 48.25 & 66.75 \\\hline
    \textbf{DeepL}  & 29.27 & 50.04 & 61.26 \\\hline
    \textbf{Custom} & 37.14 & 55.80 & 53.32 \\\hline
\end{tabular}}
\caption{Scores given by sacreBLEU for the custom and publicly available systems on \textit{Skyrim}.}
\label{table4}
\end{table}

\begin{table}[H]
\renewcommand{\arraystretch}{1.3}
\captionsetup{justification=centering}
\noindent\resizebox{\columnwidth}{!}{
\begin{tabular}{|l|c|c|c|}\cline{2-4}
    \multicolumn{1}{l|}{} & \textbf{BLEU $\Uparrow$} & \textbf{chrF2++ $\Uparrow$} & \textbf{TER $\Downarrow$} \\\hline
    \textbf{Google Translate} & 26.05 & 45.35 & 72.39 \\\hline
    \textbf{DeepL}  & 27.60 & 47.04 & 67.94 \\\hline
    \textbf{Custom} & 31.18 & 48.80 & 62.96 \\\hline
\end{tabular}}
\caption{Scores given by sacreBLEU for the custom and publicly available systems on \textit{Fallout~4}.}
\label{table5}
\end{table}
\end{multicols}

\vspace{-.8em}

These automatic metrics are heavily dependent on form and processing, but this evolution gives an idea of the improvement, which according to \cite{toralway015} should achieve at the very least a BLEU score of 20 to be useful in a post-editing workflow. To better describe the results, however, this last section offers a more qualitative analysis.

\section{Analysis}

As noted by \cite{marie2021}, MT evaluation has become less comprehensive and reliable over the last years, but one way to overcome this pitfall is to support automatic evaluation with human analysis. For this reason, we present two sets of evaluations. The first conveys the broad trends observed in our analysis of the two games presented previously, which we cannot publish for copyright reasons. We will further illustrate these remarks with a second set offering concrete examples from the translation of a fan-made mod. We believe the best way to transparently convey and judge the result of an automatically translated text is to provide a full and continuous example of this translation, so we have made it available online and we report here the index of the quoted sentence where appropriate.\footnote{~\url{https://gitlab.uliege.be/dhansen/VGMT-article/}.\\We provide 323 segments, broken down into four illustrative excerpts. They are mainly made up of the dialogues that are often raised as criticism against MT and that were chosen because they formed mostly coherent pieces of conversation, with the exception of disrupting segments that are typical of video game files. The first excerpt is singular, as it contains segments from the 2002 game, which therefore also appear in our original training data. We included it nonetheless, as these intertextual references can be frequent and we will see that the system sometimes takes liberties with these. There follows three other excerpts for which the quality can be judged respectively as good, average or less adequate. For each of these, we also provide another machine translation suggested by the generic tool \textit{DeepL} on 25/02/2022.}

The mod, named \textit{Beyond Skyrim}, comes from the fifth instalment of the Elder Scrolls series and has been in development for multiple years, with the aim of adding seven more provinces faithful to the game universe in addition to the only one featured in the original title. This includes a province featured in a former opus of the franchise, \textit{The Elder Scrolls III: Morrowind} (Bethesda Softworks, 2002), on which we focused our attention. While there remains typical MT errors that require the intervention of a trained translator, the actual texts produced by our adapted engine for this mod made it also clear that such tools could be useful to the translation of games developed by studios and fan-made content.

\subsection{General observations}

As a note, we should clarify that these are not exhaustive analyses, but rather the main trends observed when assessing the resulting translations and subsequently using this material with students to initiate discussions about MT. We also provide a few examples, but we strongly invite readers to have their own look at the output. The main and somewhat expected observation is that although \textit{Google Translate} and \textit{DeepL} reach a respectable BLEU score and produce mostly understandable solutions, the concrete results seem hardly adapted to a video game translation, in particular when compared to an adapted system.

Indeed, both online systems have a hard time when it comes to \textbf{register}, for instance. Whereas our system was generally able to render the high register and medieval feel of \textit{Skyrim}'s original French translation as well as the particularly colloquial, even crude register of \textit{Fallout}, both were neutralized with the online tools (cf. in the provided translation to segments 7, 26). We noticed nonetheless that the training data from one game contaminated translations for the other, as crude language appeared on rare occasions in \textit{Skyrim} and \textit{Fallout}'s dialogues were sometimes too polite, or the French distinction between the informal/formal 2$^{nd}$ person singular/plural was not observed (97, 98, 115). This suggests that even with an in-domain data set, it could be useful to go beyond this simple video game adaptation and give different weights to the training data in order to further adapt the system on specific video games or franchises.

Another evident advantage of a custom system for which we do not even need to point to a given example is its ability to translate virtually all of the fictional content and terms from the games, be it names of people, institutions, places, spells, monsters, items, mythological concepts, expressions from constructed languages... This is all the more salient as terms such as ``Dragonborn'' (\textit{Enfant de dragon}), ``Greybeards'' (\textit{Grises-Barbes}) or ``Skyrim'' (\textit{Bordeciel}) are at the very core of the story, and this \textbf{vocabulary} is always left untranslated by \textit{Google Translate} or \textit{DeepL}. A particularly representative example comes from \textit{Fallout 4}, where ``Dogmeat'', the name of a dog companion, has been translated by \textit{la viande de chien} in \textit{DeepL} (literally ``the canine flesh'') instead of the meat brand for dogs (\textit{Canigou}) that is used in the game. Closely related is the rules for the \textbf{capitalization} of words that are much more restrictive in French and observed by the personalized engine, whereas the online tools systematically copy the English case.

We have also noticed that our MT system has learned translation strategies used to anticipate \textbf{gender} variation. Indeed, players can often choose their gender in video games, but this is even more visible in \textit{Skyrim}, where they can also choose between various fantasy races. While this is not a problem in English, many words referring to the player character change according to gender in French. As such, the common translation strategy is to neutralize the potential gender pitfall by deleting every direct mention of gender (as illustrated above), not using tenses that require a feminine or masculine form (cf. 28), omitting gendered words (218) or using generic terms/paraphrases in their stead (16). Our custom engine has applied this strategy systematically, interestingly showing that NMT can learn not only specific vocabulary but also translation strategies that can anticipate common sources of error in video game translations.

Lastly, the adapted MT engine offers surprising results as a whole with the translation of these \textbf{dialogues}. This if even true of khajiit and argonian languages, two races in the game speaking from a specific third-person point of view. There remains, however, notable difficulties, such as with characters speaking in a way that is depicted as drunk (250--252), with some --- not all --- idioms (208), colloquial speech (117, 212), or oral language that is ripe with pauses or incomplete sentences (128, 303) that the system tries to complete by itself.

A localization-related difficulty shared by all systems lies in the \textbf{ambiguities} between imperatives and infinitives, that take the same form in English but not in French. Paired with the lack of context, it is thus very hard for MT to differentiate between dialogue choices, orders given to the player or quest objectives and maintain a coherent translation. Hence, both French forms -\textit{er} / -\textit{ez} appear somewhat randomly. This is a challenge that human translators can face if they are not given any context, but that could be alleviated for both human and machine by using tags. A final obstacle faced by MT which makes human intervention imperative is its tendency to \textbf{translate literally} (61), especially when idioms are concerned (300), although we should perhaps remind that our model was trained with extremely limited data. More resources could improve this last point, as well as those that follow.

A last and interesting observation is that MT happens to correct human errors, which can be due to either time pressure or a lack of intersemiotic context, such as when the text describes a geographically situated object in the game and the translator cannot rely on any information. These scenarios reinforce our idea that MT could improve the quality of the final translation, by offering alternative solutions or encouraging translators to reflect on ambiguities.

\vspace{-.1em}
\subsection{Machine translation of fan-made content}

Leaving behind questions of reference or comparison between generic and adapted models, we now want to delve into more a more language-specific analysis. For this, we focus exclusively on the translation of our mod and we continue to give references to specific examples.

The main observation, which is a known characteristic of NMT, is that while the system does not have many issues with form or fluency, there are many problems with meaning or adequacy. The most obvious and problematic are instances where the sentence is grammatically correct but has an \textbf{opposite meaning} (97, 112, 146, 263). In other cases, words are correctly translated but not in the given context, creating a \textbf{shift in meaning} (67, 114). On rarer occasions, words simply have a \textbf{wrong translation} (114, 132), or there is an \textbf{omission} (70) or \textbf{hallucination} (318) in the target text.

Our review has also highlighted other minor issues, for instance with \textbf{determiners}, especially if they are omitted in English (55, 75, 272). \textbf{Errors in the source text} can be problematic (289), even though this is not always the case (17). Finally, truecasing must be ignored seeing as case serves to distinguish most fantasy-related words, but terms in \textbf{all caps} usually confuse the machine (108, 244).

As a final note, we found that despite the presence of some segments in the training data for the first excerpt, the machine took some liberties with the translation. Some of these could arguably be said to work better than the original translation (36), but other dialogues show that the translation can simply vary while being equivalent to the reference (30), or introduce an error in the text (29). It is therefore necessary to remain mindful of these weaknesses. And while MT can bring more coherence between titles of the same franchise, these liberties can also be problematic, for instance with in-game books for which the translation should not be changed. With dialogues, however, these might lessen the feeling of \textit{déjà vu} for players and rejuvenate the game experience through retranslation.

\section{Discussion}

As we have established in previous sections, we were able to achieve very encouraging results on a video game translation experiment by training an MT system on a surprisingly small in-domain data set. This was both expected and unexpected, in the sense that using relevant training data would logically boost performance and allow the system to assimilate specific terms or previously translated phrases, but we had not foreseen that this custom engine would learn abstract translation strategies that are particular to this domain, such as neutralizing variables when translating from a neutral into a gender-inflected language. We therefore think that, if implemented the right way, MT might be susceptible of helping with terminology and formal or standardized expressions, to offer relevant suggestions and maybe speed up the process, becoming a useful tool for either professional translators or amateurs working on fan translations.

This requires nevertheless that studios consider the translation process and resources as an integral part of the game development, which is not often the case and a source of problems even with human translation. Indeed, localization should be planned early in the development by rigorously following internationalization steps to accommodate translations into any language \citep{chandlerdeming2012}. The use of MT further emphasizes this need, which could mean ensuring that texts are not hard-coded and mixed with lines of code, using tags that could be useful to human and machine to alleviate the ambiguities and lack of context, or refraining from imposing formal constraints to account for differences with languages other than English.

On the other hand, the video game industry benefits from a significant advantage, which is that all translations are already aligned and easily convertible into language resources. Yet, our study also highlighted the parallel between the decontextualized working conditions of the translator and the machine translation system. This lack of contextual information and resources is pointed as one of the main challenges translators have to face in the field \citep{bernalmerino2008,theroine2021}, and one might wonder in these circumstances why translators so rarely have access to some kind of translation memory, or at least an official glossary. This alignment incidentally makes it easy to train MT systems tailored to the particular video game genre, or even specific works, which our study has shown to be able to effectively reproduce the expected terminology and offer relevant suggestions.

This might even come as a welcome change, if we consider that some professionals in the field (around 12\%) already use MT in their workflow \citep{rivasginel2021}. With the convenience of a customized engine, we could even expect this number to rise, as~such~a~tool help deliver faster translations, provide otherwise lacking information to translators and, most important of all, boost creativity. This last point might be the biggest asset of human translators in this domain, given that it is required to overcome the numerous technical and linguistic challenges that are typical of video game translation \citep{diazmonton2007}, without mentioning the very creative nature of the content itself. To this end, we plan to delve into a closer reading of these \textit{ad hoc} MT solutions from the perspective of inter-semiotic dynamics \citep{houlmont2022}.

In this respect, MT could free up time for creative thinking and particularly challenging segments or simply offer relevant suggestions, especially if it is used in combination with other tools and material such as TMs, reference corpora or termbases. However, we think this would happen if MT is used not as a first draft that may constrain the translation, but as a suggested translation akin to a TM match that would help correct points of interpretation in the source text, maintain stylistic and lexical cohesion, and even spot eventual mistakes in the target text.

On the amateur side, this technology could prove even more useful to fans for the translation of their user-generated content, thus expanding the success and replayability of their favourite games, or allowing games that were never intended to be translated to reach a wider audience. This, in turn, could promote cultural exchanges and sensitivity through a medium that otherwise tends to erase such influences against players' own expectations \citep{mandiberg2015}.

All of these positive aspects of MT conversely depend on various ethical issues, the main one being the reason for its introduction. Relying on machine translation only for the sake of productivity, without human intervention, or blindly forcing its deployment even when it is not appropriate is sure to have a drastic impact on quality, creativity and the overall appeal of the game. Such habits evidently tie into much broader issues, some of which MT might even reinforce as in the case of amateur translations being used by companies seeking to cut costs. Finally, we should not forget that all of the resources used to train these systems depends on the quality of the material that is provided by human translators, whose rights seem to become less and less apparent as translation technologies progress \citep{bowker2021}. We therefore hope that this exploratory article and its conclusions will hopefully start some discussions in the video game, or the cultural sector as a whole.

\small
\bibliographystyle{amta2022}
\bibliography{amta2022}

\section*{Appendix~A. Corpus Specifications.}

\begin{table}[H]
\renewcommand{\arraystretch}{1.4}
\begin{tabularx}{1\textwidth}{|
    >{\hsize=.14\hsize\linewidth=\hsize}X|
    >{\hsize=1.65\hsize\linewidth=\hsize}X|
    >{\hsize=1.45\hsize\linewidth=\hsize}X|
    >{\hsize=.76\hsize\linewidth=\hsize\raggedleft\arraybackslash}X|}\cline{2-4}
    \multicolumn{1}{>{\hsize=.14\hsize\linewidth=\hsize}X|}{}
        & \textbf{Game}
        & \textbf{Developer}
        & \multicolumn{1}{>{\hsize=.75\hsize\linewidth=\hsize}X|}{\textbf{Size}}\\\hline
    \textbf{1.}
        & \textit{Baldur's Gate}
        & BioWare, 1998
        & 19 K segments\\\hline
    \textbf{2.}
        & \textit{Baldur's Gate II: Shadows of Amn}
        & BioWare, 2000
        & 62 K segments\\\hline
    \textbf{3.}
        & \textit{Darkest Dungeon}
        & Red Hook Studios, 2016
        & 8 K segments\\\hline
    \textbf{4.}
        & \textit{Divinity: Original Sin II}
        & Larian Studios, 2017
        & 85 K segments\\\hline
    \textbf{5.}
        & \textit{Fallout}
        & Black Isle Studios, 1997
        & 26 K segments\\\hline
    \textbf{6.}
        & \textit{Fallout 2}
        & Black Isles Studios, 1998
        & 56 K segments\\\hline
    \textbf{7.}
        & \textit{Fallout 3}
        & \mbox{Bethesda Game Studios, 2008}
        & 49 K segments\\\hline
    \textbf{8.}
        & \textit{Fallout: New Vegas}
        & \mbox{Obsidian Entertainment, 2010}
        & 64 K segments\\\hline
    \textbf{9.}
        & \textit{Fallout 4}
        & \mbox{Bethesda Game Studios, 2015}
        & \mbox{126 K segments}\\\hline
    \textbf{10.}
        & \textit{Planescape: Torment}
        & Black Isle Studios, 1999
        & 39 K segments\\\hline
    \textbf{11.}
        & \textit{Pillars of Eternity}
        & \mbox{Obsidian Entertainment, 2015}
        & 48 K segments\\\hline
    \textbf{12.}
        & \textit{Star Wars: Battlefront II}
        & Pandemic Studios, 2005
        & 4 K segments\\\hline
    \textbf{13.}
        & \textit{The Elder Scolls III: Morrowind}
        & Bethesda Softworks, 2002
        & 37 K segments\\\hline
    \textbf{14.}
        & \textit{The Elder Scrolls IV: Oblivion}
        & \mbox{Bethesda Game Studios, 2006}
        & 40 K segments\\\hline
    \textbf{15.}
        & \textit{The Elder Scrolls V: Skyrim}
        & \mbox{Bethesda Game Studios, 2011}
        & 70 K segments\\\hline
    \textbf{16.}
        & \textit{The Witcher 2: Assassins of Kings}
        & CD Projekt, 2011
        & 32 K segments\\\hline
    \textbf{17.}
        & \textit{The Witcher 3: Wild Hunt}
        & CD Projekt RED, 2015
        & 78 K segments\\\hline
    \textbf{18.}
        & \textit{Torment: Tides of Numenéra}
        & inXile Entertainment, 2017
        & 49 K segments\\\hline
    \textbf{19.}
        & \textit{Ultima VII: The Black Gate}
        & Origin Systems, 1992
        & 12 K segments\\\hline
    \textbf{20.}
        & \textit{Ultima VIII: Pagan}
        & Origin Systems, 1994
        & 6 K segments\\\hline
    \textbf{21.}
        & \textit{Ultima IX: Ascension}
        & Origin Systems, 1999
        & 9 K segments\\\hline
    \textbf{22.}
        & \textit{Wasteland 2}
        & inXile Entertainment, 2014
        & 37 K segments\\\hline
    \multicolumn{3}{|X|}{\textbf{Total}}
        & \mbox{956 K segments}\\\hline
\end{tabularx}
\label{table6}
\end{table}

\end{document}